\documentclass{article}

     \usepackage[nonatbib, final]{neurips_2019}

\usepackage[utf8]{inputenc} %
\usepackage[T1]{fontenc}    %
\usepackage{hyperref}       %
\usepackage{url}            %
\usepackage{booktabs}       %
\usepackage{amsmath}
\usepackage{amsfonts}       %
\usepackage{nicefrac}       %
\usepackage{microtype}      %

\usepackage{graphicx}
\usepackage{caption}
\usepackage{placeins}
\usepackage{xcolor}
\usepackage{bbm}
\usepackage{mathtools}

\DeclareMathOperator{\LogisiticLoss}{LogisticLoss}

\title{A Self-Supervised Auxiliary Loss for Deep RL in Partially Observable Settings}

\author{%
  Eltayeb K. E. Ahmed \\
  African Masters of Machine Intelligence\\
  African Institute for Mathematical Sciences, AIMS\\
  Kigali, Rwanda\\
  \texttt{eahmed@aimsammi.org} \\
  \And
  Luisa Zintgraf \\
 University of Oxford \\
  \texttt{luisa.zintgraf@cs.ox.ac.uk} \\
   \AND
   Christian A. Schroeder de Witt \\
   University of Oxford \\
   \texttt{cs@robots.ox.ac.uk} \\
  \And
  Nicolas Usunier \\
 Facebook AI Research \\
 \texttt{usunier@fb.com}
}

\begin{document}

\maketitle

\begin{abstract}
In this work we explore an auxiliary loss useful for reinforcement learning in environments
where strong performing agents are required to be able to navigate a spatial environment. The
auxiliary loss proposed is to minimize the classification error of a neural network classifier that predicts whether
or not a pair of states sampled from the agents current episode trajectory are in order. The classifier
takes as input a pair of states as well as the agent's memory. 
The motivation for this auxiliary loss is that there is a  strong correlation with which of a pair of states is more recent in the agents episode trajectory and which of the two states is spatially closer to the agent. 
Our hypothesis is that learning features to answer this question encourages the agent to learn and internalize in memory representations of states that facilitate spatial reasoning. 
We tested this auxiliary loss on a navigation task in a gridworld and achieved $9.6$\% increase in accumulative episode reward compared to a strong baseline approach.
\end{abstract}

\section{Introduction}

Deep reinforcement learning (RL) generally requires a lot of data before strong policies can be learned. This is true even for some of the simplest environments. This is referred to as sample inefficiency, and tackling this problem has been the goal of a lot of research carried out in deep reinforcement learning~\cite{q_learning_sample}. 
Sample inefficiency can be partially attributed to the fact that in many problems the reward signal is very sparse and provides only a small amount of information per episode. This can result in agent learning very little in episodes where no new rewards are received. This is a clear shortcoming in many model-free RL algorithms. 
This is because during interactions with the environment one should be able to learn useful information of the dynamics of the environment, i.e., how the world in which the agent is embedded actually works. 
This type of information should be learnable even from episodes where the agent does not accomplish the full task successfully and does not receive any reward.

Model-based methods can be more sample efficient since every episode contains information useful for learning the transition function of the environment, regardless of whether or not a useful reward signal was received. However, as to date, model-based approaches are yet to reach the asymptotic performance of model-free algorithms. A middle-ground stance of increasing popularity is to augment model-free algorithms with auxiliary tasks (which can be either supervised in nature, in which case we refer to them as auxiliary losses, or reinforcement learning tasks, which we refer to as auxiliary tasks). These tasks and losses are selected such that the features the network learns to solve the auxiliary tasks are also useful for learning a good policy. Auxiliary tasks can require additional supervision such as loop closure detection and depth prediction~\cite{navigate_complex}, or not require additional supervision (be self-supervised) such as predicting the reward or next state~\cite{starcraft_defogger, unreal}. 

In this work we propose a novel self-supervised auxiliary loss for partially observable Markov Decision Processes (POMDPs) which encourages the agent to reason about the temporal and spatial structure of the environment. Many navigation tasks require the agent to remember which states it has already visited, and be able to reason about recently seen states in some sense. This is to be able to form some sort of internal map of its immediate surroundings and to reason about the agents current position.
Building on this, our auxiliary loss was designed such that for the agent to be able to minimize the auxiliary loss we proposed the agent needs to reason about how recently an observation sampled from the current episode trajectory was seen. Such a task can be solved easily if the agent learns features that capture the spatial structure of the world. This is due to the fact that observations that correspond to locations spatially far away in the environment will not be in the agent's very recent past and vice versa. Training with our auxiliary loss requires no external labelling and initial results on a simple gridworld-based environment show that using this auxiliary loss yields a $9.6$\% increase in episode score when compared to a baseline approach.

\section{Related Work}
The use of additional machine learning tasks to benefit performance on a target machine learning class is becoming increasingly popular. There exist multiple approaches for incorporating features learned on one task to enhance performance on another. One approach to achieve this is multi-task learning~\cite{multitask_learning} where the additional task is learned with parameters being shared between the additional task and the main task. Another approach is transfer learning where typically a neural network is pre-trained on one task before being fine-tuned on another. This approach is behind many of the large successes in commoditizing and commercializing computer vision algorithms where initializing to pre-trained networks is now becoming a default~\cite{imagenet_tl} and not the exception. The tasks used in computer vision pre-training are typically supervised classification tasks. 

On the other hand, natural language processing is another field which recently has started to benefit from pre-training, with the fast rise of language modeling pre-trained models such as BERT~\cite{bert}. The language modeling pre-training tasks are self-supervised tasks where one tries to predict masked or hidden words given some other words in its vicinity. This allows for the model to learn about the meanings of words through observing how each word appears in different sequences, with the learning objective requiring the model to learn internal representations for words that allow predicting it's neighbours. These representations are found to capture the meanings of words quite well. 

Using similar reasoning, the reinforcement learning community has recently started exploring the use of auxiliary losses that involve predicting the next state~\cite{dae} or an embedding of the next state~\cite{deepmdp} given an embedding of the current state. Other work has pushed this idea even further training a generative model to predict multiple states into the future~\cite{generative_rl}. These goals are used as auxiliary losses during training and they encourage the model to include features relating to the dynamics of the environment in its state embeddings. These features are useful for constructing a policy as for most tasks a reasonable understanding of the dynamics of the environment is a pre-requisite for obtaining a strong policy. Training with an auxiliary loss that trains a model to predict future rewards has also been shown to help models internalize environment dynamics~\cite{unreal, deepmdp}.

Other auxiliary losses used in reinforcement learning have focused on tasks that require learning features of states that contain visually represented information. Approaches have varied, with some using supervised tasks such as depth prediction~\cite{navigate_complex} and semantic segmentation~\cite{house3d}, while others have opted for unsupervised tasks such as autoencoder based reconstruction~\cite{world_models}. 

In the domain of solving navigation related task the NeuralSLAM framework~\cite{neural_slam} suggests a custom architecture coupled with predicting the agents own location as well as the agents future position.

All the above mentioned auxiliary losses are differential quantities that can be added to the loss function and directly optimized using gradient based optimization. Other methods of providing the agent with richer training signals are based on quantities that represent rewards or pseudo-rewards which are optimized using reinforcement learning methods. In visual navigation environments maximizing the pixel difference between successive frames for example was shown to encourage the agent to explore new regions of the environment~\cite{unreal}. In the same work the authors also showed that training the agent to control features of the environment can be a useful auxiliary task to learn, and that such tasks would teach the agent how to meaningfully interact with the environment and manipulate it. Furthermore these features can be learnt features instead of handcrafted. In their work the authors made use of fact that many of activations of the neurons in later layers of the policy network correspond to meaningful environment features, hence the auxiliary task concretely would be to train the agent to manipulate the environment such as to activate or deactivate specific neurons in its network.

\section{Methodology}

In this paper we consider the setting of partially observable Markov Decision Processes (POMDPs) which are defined by a tuple $(\mathcal{S}, \mathcal{A}, \mathcal{O}, T, U, R)$. $\mathcal{S}$ is the state space, $\mathcal{A}$ is the action space and $\mathcal{O}$ is the observation space. We denote $s_t \in \mathcal{S}$ as the latent state in time t. When an action $a_t \in \mathcal{A}$ is carried out the state changes according to a distribution defined by the transition function $s_{t+1} \sim T(s_{t+1} | s_t, a_t)$. After this transition takes place the agent receives a noisy or partially occluded observation $o_{t+1} \in \mathcal{O}$ according the distribution $o_{t+1} \sim U(o_{t+1}| s_{t+1}, a_t)$ and a reward $r_{t+1} \sim R(r_{t+1} | s_{t+1}, a_t)$.
The agent's goal is to learn a policy $\pi$ that maximizes expected return, with the return being the expected discounted reward 
\begin{equation}
	J = E_{p(\tau)} \left[ \sum_{t=1}^{T} \gamma^{t-1} r_t \right] .
\end{equation}
The expectation here is taken over the distribution $p$ over trajectories $\tau$ induced by the policy, and $0 \le \gamma < 1$ is a discount factor.

A common approach to solving POMDPs using deep reinforcement learning is to represent the policy as a recurrent neural network $\pi_\theta(a_t | o_t, h_t)$ parameterized by $\theta$, where $o_t$ is the current observation, and $h_t$ is the hidden state of the recurrent part of the network \cite{drqn}. At the beginning of an episode, the hidden state is initialized to zero, $h_0 = \mathbf{0}\in\mathbb{R}^{N_h}$ where $N_h$ is the size of the hidden state. 

We want to define our auxiliary loss such that it encourages the agent to represent spatial or temporal structures in its hidden state $h_t$. In particular, we propose to learn a binary classifier $f$ which can predict whether two observations the agent has already seen, $o_i$ and $o_j$ ($i\not=j$ and $i, j \le t$), are in order or not:
\begin{equation}
f(h_t, o_i, o_j) = 
\begin{cases}
+1 ~~ \text{ if } ~~ i<j \\
-1 ~~ \text{ if } ~~ i>j 
\end{cases} .
\end{equation}

We parameterize this classifier with a deep neural network and train this classifier using a logistic loss function on the output of the binary classifier optimizing both the parameters of the classifier and the layers of the agent's network that are involved in computing the agents hidden state vector. To train the agents we used the Synchronous Advantage Actor-Critic Algorithm (A2C) which is a synchronous version of the A3C algorithm \cite{a3c}. We also used entropy regularization for exploration. We added the loss of the above defined classifier to the loss of the reinforcement learning algorithm after scaling the loss with a weight $\beta$, where $\beta$ is a hyper-parameter to be tuned. 

The following equations define the overall loss function used:
\begin{align}
\mathcal{L} (\theta, \phi) &= \log \pi_\theta (a_t | o_t, h_t) \left(r_{t+1} + \gamma V_\phi (o_{t+1} | h_{t+1}) - V_\phi(o_t | h_t)\right)     \notag\\
&\phantom{{}=1} + \alpha \mathcal{H}\left(\pi_\theta(a_t | o_t, h_t) \right) \\
&\phantom{{}=1} + \beta \sum_{\mathclap{(o_i, o_j) \in \mathcal{S}}} \LogisiticLoss \left( f(h_t, o_i, o_j), \mathbbm{1}_{i < j}  \right) \notag ,
\end{align}

where $\gamma$ is the discount factor, $V_\phi$ is an estimate of the value function parameterized by $\phi$, $\mathcal{H}\left(\pi_\theta(a_t | o_t, h_t) \right)$ is the exploration encouraging entropy term (and this term is weighted by $\alpha$), and $\mathcal{S}$ is the set of examples sampled at each timestep to train the auxiliary loss classifier .

\begin{figure}[t]
\centering
	\includegraphics[width=\textwidth]{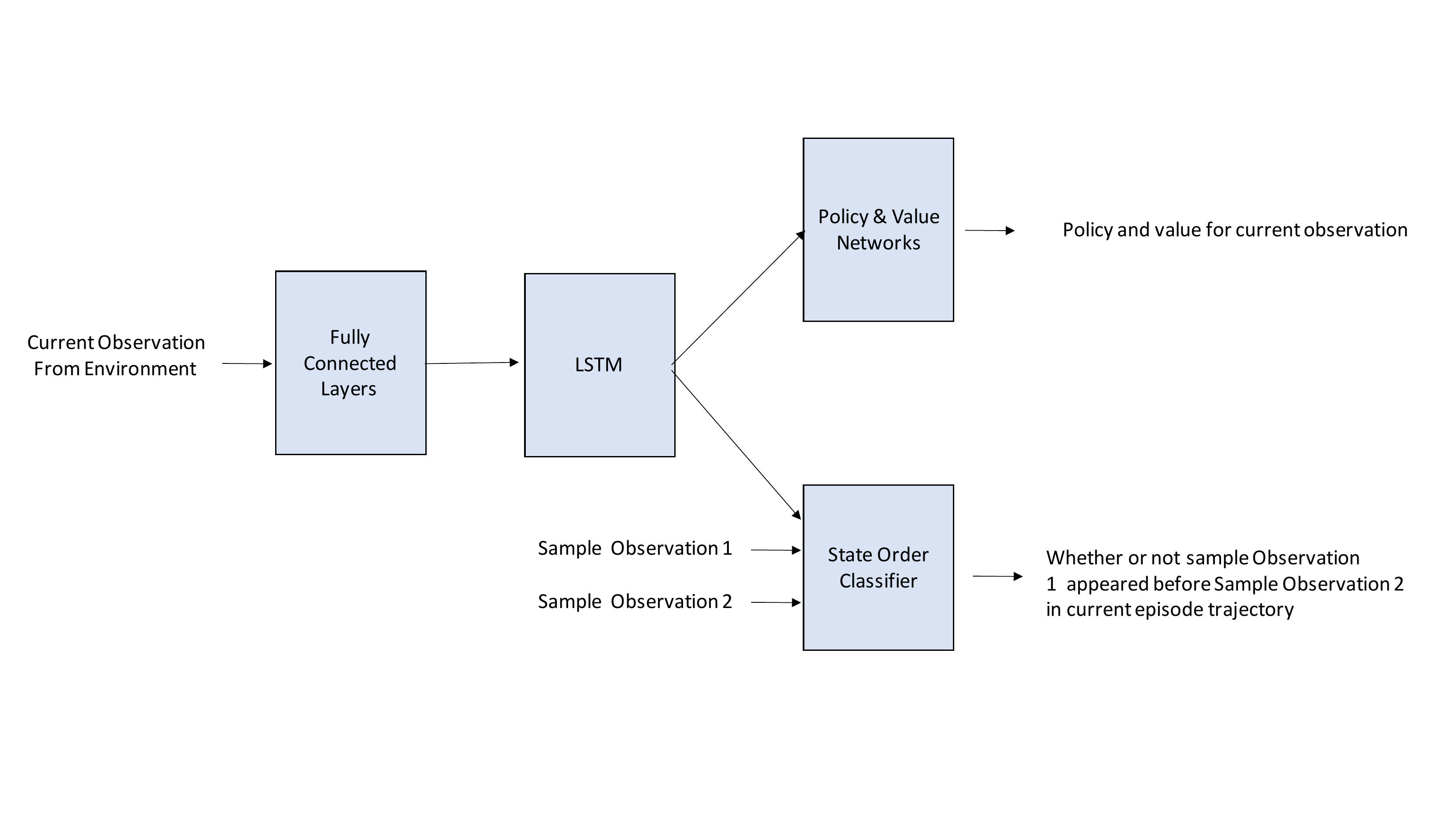}
	\caption{This diagram shows the forward flow of information to compute value, policy and auxiliary loss. An Advantage Actor-Critic loss was computed from the output of the policy and value networks while a classification loss was computed from the output head of the state order classifier. Both losses were back-propagated to train the network parameters.}
	\label{fig:arch}
\end{figure}

To compute the auxiliary loss at each time step a number of positive examples $(o_i, o_j)$ with $o_i$ preceding $o_j$ in order of appearance is sampled from the episode trajectory, a negative batch of $(o_i, o_j)$ with $o_j$ preceding $o_i$ are also sampled. Sampling $o_i$ and $o_j$ from only the $k$ most recent observations instead of the entire episode trajectory lead to slightly improved results (with $k$ being another hyperparameter to tune). A diagram showing the forward flow of information is shown in Figure~\ref{fig:arch}.

The motivation for this auxiliary loss is that once the agent learns to move in a slightly purposeful manner through the environment, states appearing later in the episode trajectory are closer to the agent's current position than states appearing earlier. Learning to answer queries relating to the distance of states requires having spatially informative representations of states as well as a representation of the agent's current position in memory. This is useful because it is a pre-requisite for doing spatial reasoning. In environments with navigation components agents with strong spatial reasoning capabilities have a distinct advantage over agents without.

\section{Empirical Evaluation}

\begin{figure}[t]
	 \centering
	\includegraphics[width=100px]{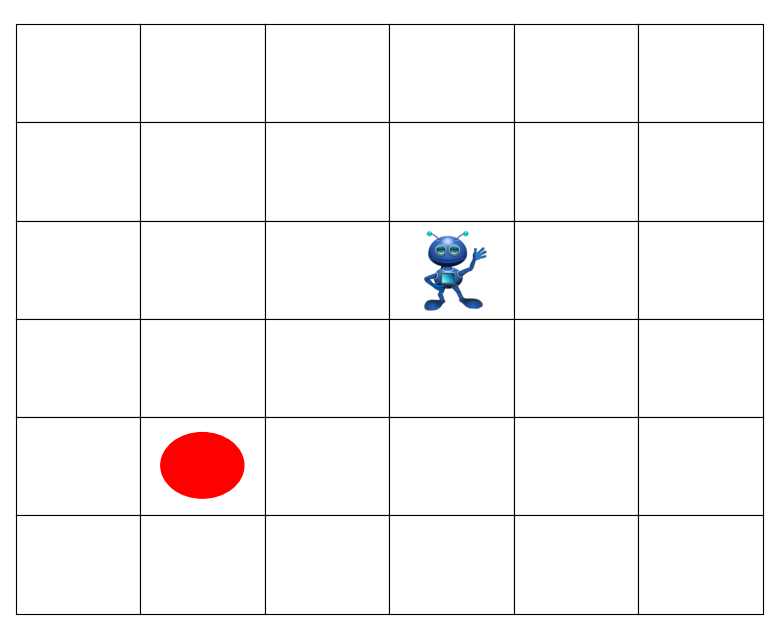}
	\caption{This is a visualization of the gridworld environment the auxiliary loss was tested on. The blue robot represents the agent and the red circle represents the goal. The agent can only see the content of the square it is currently on and its coordinates. The agent needs to search for the goal in order to obtain a reward.}
	\label{fig:env}
\end{figure}

In this work we study agents navigating in a partially observable $6\times6$ gridworld. At every time-step the agent position corresponds to a single cell in the grid. At every time-step the agent observes the coordinates of its current location as well as whether or not the square it is at contains the goal. The actions that the agent can take at each time-step is to move one square in one of the four cardinal directions or to not move at all. The agent receives a reward of $+1$ if the agent is on the goal and zero otherwise. At the start of every episode both the agent and the goal are spawned on random squares. The episodes have a fixed length of $50$ time-steps. A visualization of the environment is shown in Figure~\ref{fig:env}.

To perform well in this task the agent needs to learn to search the space in an effective manner. In order to do this the agent needs to remember which areas it has already explored, in order to not waste time re-searching places it has already been. 

The algorithm used was A$2$C with $32$ worker threads. The discount factor $\gamma$ was set to $0.95$. The architecture of the policy network was a number of fully connected layers followed by an LSTM which generated a summary of the observation history. This summary was used as input for a value network, a policy network and the auxiliary loss state order classifier network.

To control for noise due to the stochasticity in the algorithms we trained $10$ agents augmented with the auxiliary loss. As a baseline we used identical agents trained with the same procedure without the auxiliary loss. The baseline algorithm were also trained from $10$ different random seeds to control for noise. Figure~\ref{fig:scores} shows the change in episode score over time during training, while Table~\ref{table:scores} shows summary statistics of the after-training performance of the $10$ agents for each algorithm. From the table we can see the proposed auxiliary loss leads to a $9.6\%$ improvement in final performance. 

From the training curves in Figure~\ref{fig:scores} we see that the auxiliary loss has no effect during the beginning of training and only starts to have a positive effect after a significant portion of training. This supports the hypothesis that the loss is most useful once the agent starts moving purposely through the environment, for it is only then answering the question of which of a pair of observations was seen most recently corresponds to answering the question of which of the observations is spatially closer to the agent and conversely, at the beginning of training when the agent is moving in a totally random fashion answering the question of which of a pair of states was seen most recently does not concretely correspond to which of the states is spatially closer, since the agent is not moving in a specific direction with purpose, but is simply following a random walk.

\section{Conclusion}
In this work we proposed an auxiliary task to benefit deep reinforcement learning in partially observable settings. We proposed an auxiliary loss which is to minimize the classification error of a model that predicts from the policy networks memory which of a pair of states sampled from the agents recent pass is more recent.  The auxiliary loss suggested in this work was designed to augment spatial reasoning capabilities, which we want to explore further in future work. We tested this loss on a simple gridworld task and from these results we see that at the cost of no further supervision or labeling we were able to get a significant improvement in the performance of the agent. Future work will involve analyzing the effect of the auxiliary task on the agent's behaviour, and scaling our experiments to larger and more complex environments, as well as experimenting with other sequence derived auxiliary losses.

\begin{table}[b]
    \centering
    \begin{minipage}{0.48\textwidth}
    \centering
	\includegraphics[width=\columnwidth]{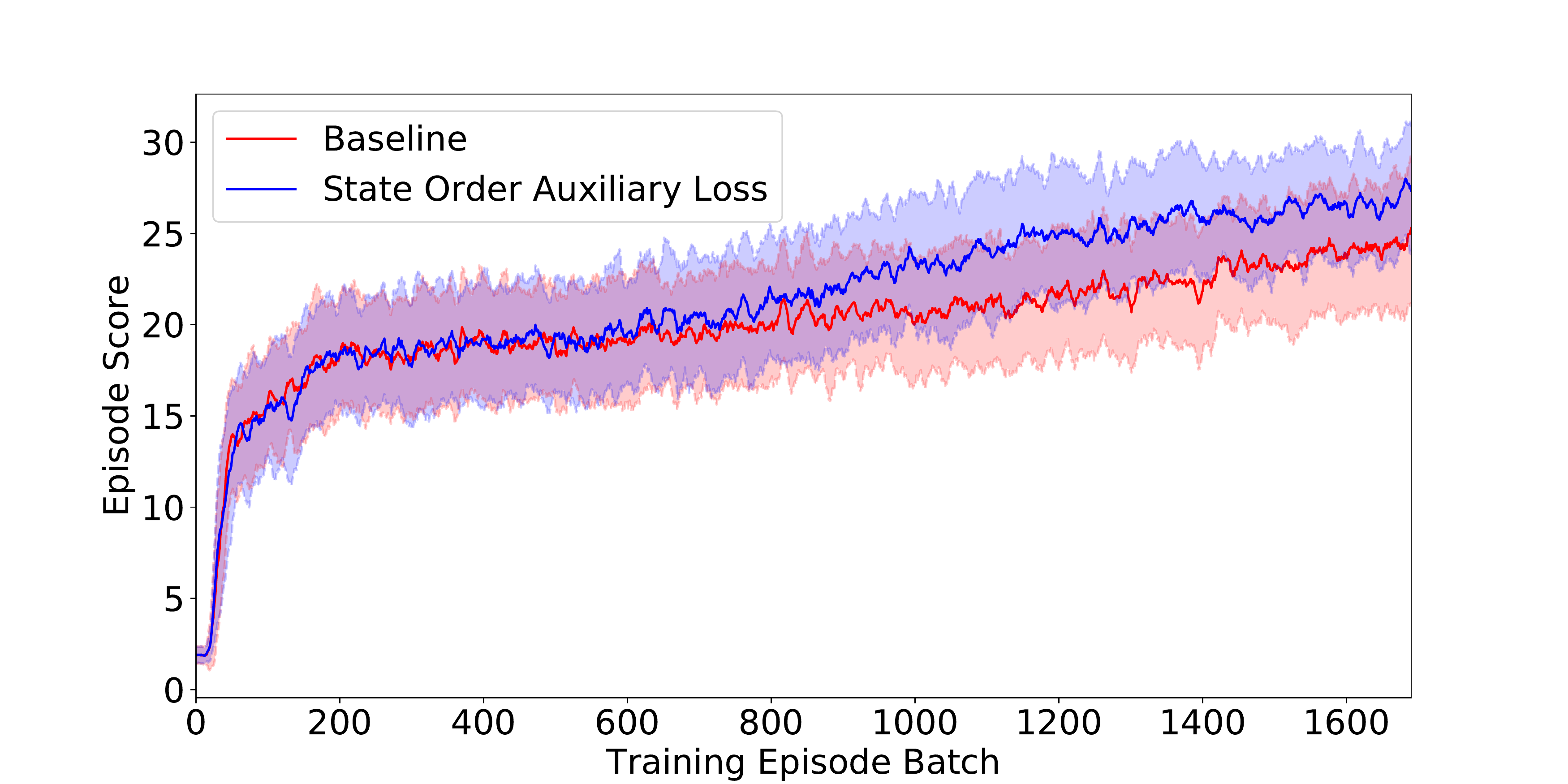}
	\captionof{figure}{Average return per episode (across ten runs) for the baseline A$2$C agent and a model with our suggested auxiliary loss. Results are averaged across 10 random seeds and the shaded area shows one standard deviation.}
	\label{fig:scores}
	\end{minipage}
	\hfill
    \begin{minipage}{0.48\textwidth}

	\centering
	\scalebox{0.8}{%
		\begin{tabular}{lll}
			\toprule
			Model & Mean & Std \\
			\midrule
			Baseline & 24.76 & 1.98\\
			State Order Auxiliary Loss Agent & 27.16 & 1.23\\
			\bottomrule
	\end{tabular}}
	\vspace{4.5em}
	\captionof{table}{Summary statistics of test evaluation for ten runs of both the baseline and an agent augmented with our auxiliary loss.}
	\label{table:scores}
\end{minipage}
\end{table}

\FloatBarrier

\bibliography{bib} 
\bibliographystyle{ieeetr}
\end{document}